\documentclass[conference]{IEEEtran}
\IEEEoverridecommandlockouts
\usepackage{cite}
\usepackage{booktabs}
\usepackage{amsmath,amssymb,amsfonts}
\usepackage{algorithmic}
\usepackage{graphicx}
\usepackage{textcomp}
\usepackage{xcolor}
\usepackage{algorithm}
\def\BibTeX{{\rm B\kern-.05em{\sc i\kern-.025em b}\kern-.08em
    T\kern-.1667em\lower.7ex\hbox{E}\kern-.125emX}}
\begin{document}

\title{Online Policy Distillation with Decision-Attention}

\author{\IEEEauthorblockN{Xinqiang Yu$^{1,2}$, Chuanguang Yang$^{1*}$\thanks{$^{*}$Corresponding author}, Chengqing Yu$^{1,2}$, Libo Huang$^{1*}$, Zhulin An$^{1*}$, Yongjun Xu$^{1}$}
\IEEEauthorblockA{\textit{$^{1}$Institute of Computing Technology, Chinese Academy of Sciences} \\
\textit{$^{2}$University of Chinese Academy of Sciences} \\
Beijing, China \\
\{yuxinqiang21s, yangchuanguang, yuchengqing22b, huanglibo, anzhulin, xyj\}@ict.ac.cn
}}
\maketitle
\begin{abstract}
Policy Distillation (PD) has become an effective method to improve deep reinforcement learning tasks. The core idea of PD is to distill policy knowledge from a teacher agent to a student agent.
However, the teacher-student framework requires a well-trained teacher model which is computationally expensive.
 In the light of online knowledge distillation, we study the knowledge transfer between different policies that can learn diverse knowledge from the same environment.
 In this work, we propose Online Policy Distillation (OPD) with Decision-Attention (DA), an online learning framework in which different policies operate in the same environment to learn different perspectives of the environment and transfer knowledge to each other to obtain better performance together.
With the absence of a well-performance teacher policy, the group-derived targets play a key role in transferring group knowledge to each student policy.
However, naive aggregation functions tend to cause student policies quickly homogenize.
To address the challenge, we introduce the Decision-Attention module to the online policies distillation framework. 
The Decision-Attention module can generate a distinct set of weights for each policy to measure the importance of group members.
We use the Atari platform for experiments with various reinforcement learning algorithms, including PPO and DQN. In different tasks, our method can perform better than an independent training policy on both PPO and DQN algorithms. This suggests that our OPD-DA can transfer knowledge between different policies well and help agents obtain more rewards.
\end{abstract}


\section{Introduction}
Reinforcement learning (RL), especially deep reinforcement learning combined with deep learning, has achieved great success in different fields, such as robotics and games \cite{li2017deep}. Deep Reinforcement Learning (DRL) is particularly effective in handling complex, high-dimensional environments, which are difficult in traditional RL methods. However, it usually needs too many iterations with the environment to make better decisions. 
Thus, how to make policy to achieve better results under the same number of iterations with the environment is very important.
Recently, Knowledge Distillation (KD) proposed a teacher-student framework, which transfers knowledge from one well-trained teacher policy to a student policy.
Policy distillation is a simple but effective method, which uses supervised learning to train student policy to align the output distribution of teacher policy. However, this method needs a pre-trained well-performed teacher policy, which is computationally expensive. Moreover, using this framework to train a student policy will cause the performance of the student policy to be limited by the performance of the teacher policy.

Similar to real life, students work together and make progress together. Online learning is a simple but effective way to improve the generalization ability of a model by training collaboratively with other models. Compared with knowledge distillation with a pre-trained teacher model, Online Knowledge Distillation (OKD) can even help student models achieve better performance. Based on this idea, we study online learning on student policies without the use of pre-trained teacher policy and the method of transferring knowledge between different policies.

In this paper, we introduce Online Policy Distillation with Decision-Attention (OPD-DA), an online learning framework in which three policies operate in the same environment and transfer knowledge to each other to make themselves obtain better performance together. 
In OKD, the simple and intuitive method is to make a student model learn from another model's output. Thus, we can introduce this method into online policy distillation and improve agent decision-making ability by making an agent learn from others' output \cite{rusu2015policy} \cite{lai2020dual}.
However, this response-based OKD method can only give limited supervisory information.
Beside this response-based OKD method, some work focuses on generating supervisory information by intermediate layer feature maps \cite{romero2014fitnets}.
We can also realize knowledge transfer through intermediate layer feature maps of different policies and improve policy decision-making ability.
Thus, in our method, we regard both the response and intermediate layer feature maps of policies as supervisory signals, during online policy distillation. 

In online policy distillation, the quality of decisions varies across different policies. Traditional OKD method that assigns the same weights to different models \cite{yang2023categories} can't produce high-quality decisions and make these policies converge well. 
We should assign individual weights to all the peers during aggregation to derive their own target distributions. 
Thus, we introduce an attention mechanism to our method, generate a distinct set of weights for each policy to measure the importance of group members, and make them perform better.
In this paper, we propose an online policy distillation framework and Decision-Attention module. We conduct experiments on the Atari game platform using PPO and DQN algorithms. The results show that our OPD-DA can achieve better performance than the independent training agent.
Our main contributions can be summarized as follows:
\begin{itemize}
  \item [1)]
    We propose an online policy distillation framework, which can make a policy learn from the decisions and intermediate layer feature maps of other policies in an online manner.
  \item [2)]
   We design the Decision-Attention module, which can highlight differences between the decisions of various policies, and integrate decisions better.
   \item[3)]
   We make experiments by using PPO and DQN algorithms to solve different tasks on the Atari game platform. And, our method performs better on all tasks.
\end{itemize}
\section{RELATED WORKS}

\subsection{Reinforcement Learning}
Unlike supervised learning and unsupervised learning, Reinforcement Learning 
make agents learn to give better decisions by interacting with their environments and obtaining rewards \cite{lu2019artificial} \cite{liu2020new}. And RL is widely used in many fields, such as games.
The development of deep learning has had an important impact in many fields, such as object detection, speech recognition \cite{radford2023robust}, and language translation \cite{dong2021survey}. The most important property of deep learning is automatically learning efficient representations of complex data. Therefore, deep learning has similarly accelerated progress in RL, when using it within RL. This is defining the field of Deep Reinforcement Learning (DRL), which can help RL performs better in more complex environments.
Researchers have proposed several commonly used DRL algorithms, such as Deep-Q-Network (DQN) \cite{mnih2013playing} and Proximal Policy Optimization (PPO) \cite{schulman2017proximal} . 



\subsection{Online Knowledge Distillation.} The seminal Online Knowledge Distillation (OKD) method dubbed Deep Mutual Learning (DML) \cite{zhang2018deep} transfers knowledge between student models during the online training and achieves better performance than independent training. Inspired by this insight, AHBF-OKD \cite{Gong_Lin_Zhang_Shen_Li_Qiao_Ren_Li_Yu_Ma_2023} designs hierarchical branches and adaptive hierarchy-branch fusion module to aggregate complementary knowledge.
OKDDip \cite{chen2020online} alleviates
the homogenization problem in previous ONE \cite{zhu2018knowledge} method by
introducing two-level distillation and self-attention mechanism. 
MCL \cite{yang2021multi,yang2022mutual,yang2023online} present a mutual constrative learning framework for online knowledge distillation.
These above methods often perform OKD by utilizing the probabilistic output. But these methods generate supervisory information by different ways.
We introduce OKD into reinforcement learning and build an online policy distillation framework.

\subsection{Response-based KD.}
Within OKD approaches, there are different forms of knowledge definition, such as response-based knowledge and feature-based knowledge.
Response-based KD focuses on learning knowledge from the last layer as a response. And it aims to align the final soft label between the teacher and student model \cite{huang2024kfc}. Response-based methods just only utilize the outcome of models and need not modify model architecture. The seminal KD is from Hinton. Its core idea is to make softened softmax from teacher and student models more similar \cite{hinton2015distilling}. For the classification task, the soft probability distribution $p$ can be formulated as Equation 1.
\begin{equation}
    p(z_{i};T) = \frac{exp(z_{i}/T)}{\sum_{j=1}^{N}exp(z_{j}/T)}
\end{equation}
where $z_{i}$ is the logit value of the i-th classes, $N$ is the number of classes and $T$ is a temperature parameter to adjust the smoothness of class probability distribution. And response-based KD can be formulated as Equation 2 \cite{yang2021hierarchical}.
\begin{equation}
    L_{res\_kd}(p(z_{S};T), p(z_{T};T)) = L_{dis}(p(z_{S};T), p(z_{T};T))
\end{equation}
where $L_{dis}$ is a distance function, such as Kullback-Leibler divergence loss.

TAKD \cite{mirzadeh2020improved} introduces an intermediate-sized model as a teacher assistant and performs a sequential KD process. PAD \cite{zhang2020prime} proposes an adaptive sample weighting mechanism with data uncertainty based on the observation that hard instances may be intractable for KD. MixSKD~\cite{yang2022mixskd} introduces image mixture and regularizes linear probability distillation.

\subsection{Feature-based KD}
Response-based KD just regards the soft label of the model as a supervision signal. However, these methods neglect some vital supervision signals from the intermediate layer features. To overcome this defect, some work focuses on using intermediate layer features \cite{huang2024etag} to supervise the training of student models, such as feature maps. The loss function of feature-based KD can be formulated as Equation 3.
\begin{equation}
    L_{fea\_kd}(F^{S}, F^{T}) = L_{dis}(\theta^{S}(F^{S}), \theta^{T}(F^{T}) )
\end{equation}
Where $F^{S}$ and $F^{T}$ represent intermediate layer features from student and teacher models. $\theta^{S}$ and $\theta^{T}$ represent the transformation function of feature maps from student and teacher models, such as attention mechanism and probability distribution. $L_{dis}$ is a distance function that measures the similarity of features from the student and teacher model. For example, Squared Error loss and Kullback-Leibler loss can both be regarded as $L_{dis}$.

FitNet \cite{romero2014fitnets} is the first work to use feature information in KD, which straightforwardly aligns feature maps from intermediate layers in a layer-by-layer manner. This simple and intuitive method can only use limited knowledge of hidden layer features. Subsequent work attempted to utilize more meaningful information from raw feature maps, which can make feature-based KD perform better. NST \cite{huang2017like} extracts activation heatmaps as neuron selectivity for transfer. HSAKD \cite{yang2021hierarchical,yang2022knowledge} introduces auxiliary classifiers to intermediate feature maps supervised by extra tasks to produce informative probability distributions. CIRKD \cite{yang2022cross} distills cross-image relational features for efficient semantic segmentation.

\subsection{Attention}
The attention mechanism in deep learning is a technique that enables a neural network to focus on important parts of the information when processing data. The concept was originally inspired by human visual attention, how our brains prioritize the most relevant parts of sensory input. The introduction of the attention mechanism has greatly improved the performance of deep learning models in various tasks \cite{yu2023dsformer,yu2024ginar}, especially in the fields of natural language processing (NLP) \cite{vaswani2017attention} and computer vision \cite{han2022survey,yang2020gated}.
The core idea of attention is to focus limited attention on key information, thereby saving resources and quickly obtaining the most effective information \cite{vaswani2017attention}. It has few parameters, high speed, parallelism, and good effect. As Bert \cite{devlin2018bert} and GPT \cite{radford2018improving} become more and more widely used, the power of the attention mechanism has been proven. 

In a standard self-attention mechanism, elements of a sequence calculate their attention scores for other elements in the sequence. In contrast, the cross-attention mechanism allows the model to calculate attention scores on one sequence, and these scores are based on information from a different sequence.
Based above, we introduce cross-attention mechanism into our method \cite{li2023knowledge} and propose Decision-Attention to capture differences between various decisions.

\section{METHODOLOGY}
\begin{figure*}
    \centering
    \includegraphics[width=10.0cm]{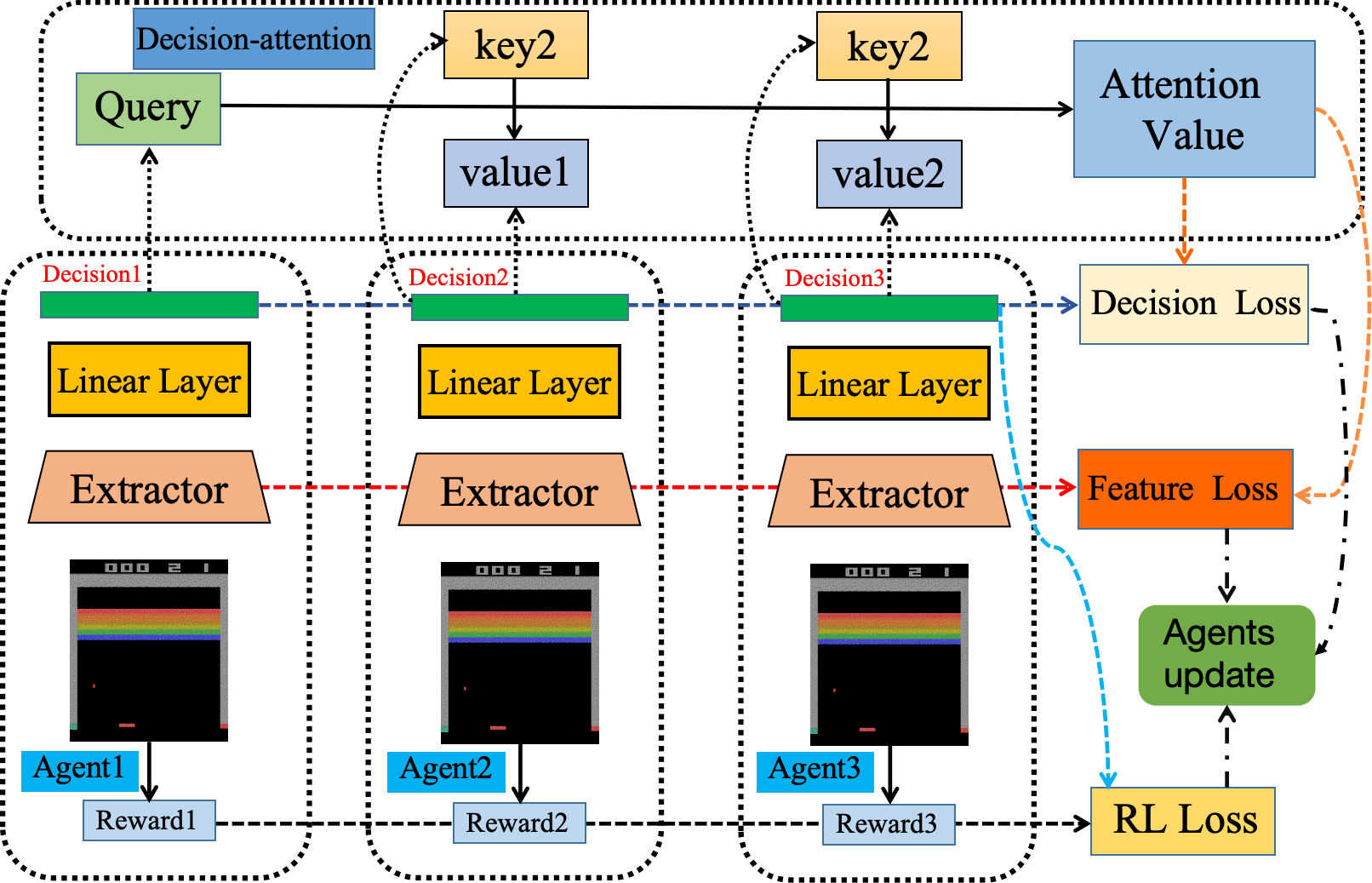}
    \caption{Overview of our method. During policy distillation, there exist three policies that include feature extractor and linear layer. Our method uses attention values to get decision loss and feature loss. These policies are trained using RL loss, decision loss, and feature loss.  }
    \label{fig:enter-label}
\end{figure*}

We propose online policy distillation framework and Decision-Attention module. During the online distillation process, these policies learn from each other by intermediate layer features and response of policies.
\subsection{Online Policy Distillation Framework}
As depicted in Figure 1, during training, T peer policies $\left\{ A_{m}(s) \right\}_{m=1}^{T}$ participate in the process of online distillation. Each policy includes a CNN feature extractor and linear layers outputting the expected rewards of each action. For RL, we aim to train policies to make decisions and obtain rewards. 
Thus, we treat the expected rewards outputted from each policy as the supervision signal and align a policy to the weighted sum of expected rewards from other policies. Given a state $s_{i}$, the expected rewards of policies $\left\{ A_{m}(s) \right\}_{m=1}^{T}$ are denoted as $\left\{ O^{i}_{m} \right\}_{m=1}^{T}$. In our OPD-DA method, in addition to using RL loss for optimization, we propose decision loss \cite{hinton2015distilling}. Using this loss, given a state, our method makes a policy align weighted sum of expected rewards from other policies.

Besides regarding expected rewards of policies as a supervisory signal, our method also uses the intermediate layer features of policies to provide more supervisory information. Given a state $s_{i}$, the intermediate layer features of policies $\left\{ A_{m}(s) \right\}_{m=1}^{T}$ are denoted as $\left\{ F^{i}_{m} \right\}_{m=1}^{T}$. Our method can help policies transfer more knowledge by using this feature-based loss function.
Using this loss, given a state, our method makes a policy align weighted sum of intermediate layer features from other policies.

 At the training stage, we jointly optimize policies $\left\{ A_{m}(s) \right\}_{m=1}^{T}$ by RL loss, decision loss, and feature loss. At the test stage, we also test policies $\left\{ A_{m}(s) \right\}_{m=1}^{T}$, jointly. 

\subsection{Learning objects}
\textbf{Learning from RL Loss.}  In RL, there are four main elements, including environment, state, action, and reward \cite{wiering2012reinforcement}. In RL, an agent performs actions in the environment according to the current state. The environment returns the next state and a reward value based on the agent's actions. 
Thus, in the training process of RL, an agent judges whether the action is good or bad and updates its policy according to the obtained rewards. For example, PPO includes an actor and critic, and their loss function is $L_{actor}$ and $L_{critic}$ in Equation 4 and Equation 5.
\begin{equation}
    L_{actor} = clip\_ratio * A^{ \pi_{\theta }}(s_{t}, a_{t})
\end{equation}
\begin{equation}
    L_{critic} = (V_{\phi}(s_{t}) - \hat{R}_{t})^{2}
\end{equation} 
And in DQN, the loss function is Equation 6.
\begin{equation}
    L = E\left\{(r + \gamma max_{a^{'}}Q(s^{'}, a^{'})-Q(s,a) ) ^{2} \right\}
\end{equation}
Where $clip\_ratio$ is clipped ratio of old and new action probabilities, $A$ is advantage value, $V_{\phi}(s_{t}) $ is new value of state $s_{t}$, and $\hat{R}_{t}$ is old value of state $s_{t}$. $Q(s,a)$ is Q-value, $\gamma$ is discount factor, and $r$ is reward. 

Thus, we use expected rewards given by actors and critics to train policies for PPO and use Q-values to train policies for DQN.

\textbf{Learning from Decision Loss.} We use the Decision-Attention module to assign weights to different decisions of policies, and construct supervision signals $O_{A}^{tea}$, $O_{C}^{tea}$, $O_{Q}^{tea}$ by ensembling decisions of other policies \cite{fukuda2017efficient}. After getting these weights from the Decision-Attention module, for PPO, OPD-DA integrates actions' expected rewards from actors and integrates states' expected rewards from critics, using Equation 7. For DQN, OPD-DA integrates actions' expected rewards from the Q-network, using Equation 8.
\begin{equation}
    O_{A}^{Tea} = \sum_{i}^{T-1} W_{A}^{i} O_{A}^{i}, \quad O_{C}^{Tea} = \sum_{i}^{T-1} W_{C}^{i} O_{C}^{i}
\end{equation}
\begin{equation}
    O_{Q}^{Tea} = \sum_{i}^{T-1} W_{Q}^{i} O_{Q}^{i}
\end{equation}
Where $T$ is the number of policies, $A$ is the actor of PPO, $C$ is the critic of PPO, $Q$ is DQN, and $W$ is the weights.

We consider transferring decisions from other policies to a policy. For PPO, KL divergence is thus used for aligning the expected rewards of each action outputted from actors using Equation 9. Mean squared error loss is used for aligning the expected rewards of states outputted from critics using Equation 10. And, KL divergence is also used for aligning the expected rewards of each action outputted from DQN in Equation 11.
\begin{equation}
    L_{kl} = KL(O_{A}^{Tea}, O_{A}^{i}) = \sum_{p=1}^{P}O_{A}^{Tea, p}(a|s)log\frac{O_{A}^{Tea, p}(a|s)}{O_{A}^{i, p}(a|s)}
\end{equation}
Where $P$ is the number of actions.
\begin{equation}
   L_{mse} = MSE(O_{C}^{Tea}, O_{C}^{i}) = (O_{C}^{Tea} - O_{C}^{i})^{2}
\end{equation}
\begin{equation}
    L_{kl} = KL(O_{Q}^{Tea}, O_{Q}^{i}) = \sum_{p=1}^{P}O_{Q}^{Tea, p}(a|s)log\frac{O_{Q}^{Tea, p}(a|s)}{O_{Q}^{i, p}(a|s)}
\end{equation}
\textbf{Learning from Feature Loss.}
 Our method also regards the feature from the last layer of the CNN feature extractor as supervisory information.
 For PPO, OPD-DA integrates intermediate features from both actors and critics, using Equation 12. For DQN, OPD-DA integrates intermediate features from the Q-network, using Equation 13.
 \begin{equation}
    F_{A}^{Tea} = \sum_{i}^{T-1} W_{A}^{i} F_{A}^{i}, \quad F_{C}^{Tea} = \sum_{i}^{T-1} W_{C}^{i} F_{C}^{i}
\end{equation}
 \begin{equation}
    F_{Q}^{Tea} = \sum_{i}^{T-1} W_{Q}^{i} F_{Q}^{i}
\end{equation}
Where $T$ is number of policies, $A$ is the actor of PPO, $C$ is the critic of PPO, $Q$ is DQN, and $W$ is the weights.

Mean squared error is used for aligning the intermediate feature of one policy with others. For PPO, our method aligns intermediate features from both actor and critic, using Equations 14 and 15. For DQN, our method aligns intermediate features from the Q-network, using Equation 16.
\begin{equation}
   L_{mse} = MSE(F_{A}^{Tea}, F_{A}^{i}) = (F_{A}^{Tea} - F_{A}^{i})^{2}
\end{equation}
\begin{equation}
   L_{mse} = MSE(F_{C}^{Tea}, F_{C}^{i}) = (F_{C}^{Tea} - F_{C}^{i})^{2}
\end{equation}
\begin{equation}
   L_{mse} = MSE(F_{Q}^{Tea}, F_{Q}^{i}) = (F_{Q}^{Tea} - F_{Q}^{i})^{2}
\end{equation}
\subsection{Decision-Attention}

In RL, a policy earns rewards by making decisions and interacting with the environment. Thus, how to make a better decision is very important for policies. For example, in DQN algorithm, the Q-network outputs Q-values for each action. And, in PPO, the actor outputs the expected rewards of each action, and the critic outputs the expected rewards of the current state. 
Then, according to these outputs, policies make decisions \cite{sutton2018reinforcement}. Therefore, in our framework, we make each policy learn the decisions of others by aligning these outputs.
Due to the independent updating of each policy, for the same states, the decisions produced by them are different. To learn the decisions of other policies better, our method should make mixed decisions based on the differences between decisions.

Our framework proposes the Decision-Attention module, which uses an attention mechanism to capture decision differences between policies. In the context of the attention mechanism commonly used in deep learning, Query, Key, and Value refer to the three important types of matrices or vectors. They have the same dimension $d_{k}$. The attention values are calculated by Equation 17.
\begin{equation}
    W = Softmax(\frac{ QK }{\sqrt{d_{k}}})V
\end{equation}
Where $W$ represents the attention values.

Our method designs the actions' expected rewards of a policy as Query, and the actions' expected rewards of other policies as Key and Value. Then, the Decision-Attention module outputs the weights of each policy. Then our method uses these weights to integrate decisions and generate a supervision signal using Equation 7 and Equation 8.

We conduct experiments on DQN and PPO algorithms. For DQN, the output of the Q-network is $\left\{ O_{Q}^{i} \right\}_{i = 1}^{T}$.  For DQN, our Attention-Decision module is formulated as Equation 18.
\begin{equation}
W_{Q}^{j} = softmax(\frac{ O_{Q}^{i}  O_{Q}^{j}  }{\sqrt{d_{k}}})O_{Q}^{j}
\end{equation}
Where $j \in [1,T]$ and $ j \neq i$, $T$ is the number of policies and $W_{Q}^{j}$ is the weight assigned to $O_{Q}^{j}$.

PPO algorithm always includes actor and critic, in which the output of actor $\left\{ O_{A}^{i} \right\}_{i = 1}^{T}$ is the expected rewards of each action and the output of critic $\left\{ O_{C}^{i} \right\}_{i = 1}^{T}$ is the expected rewards of the current state. Thus, for PPO, our Attention-Decision module is formulated as Equation 19.
\begin{equation}
    W_{AC}^{j} = softmax(\frac{ O_{A}^{i} O_{A}^{j} }{\sqrt{d_{k}}})O_{A}^{j}
\end{equation}
Where $j \in [1,T]$ and $ j \neq i$, $T$ is the number of policies and $W_{AC}^{j}$ is the weights assigned to $O_{A}^{j}$ and $O_{C}^{j}$.

In our OPD-DA, we input the same state of a task to all policies and get different expected rewards from these policies. 
Each policy learns from others using the above Decision-Attention module and updates iteratively. And in our OPD-DA method, the loss function is formulated as Equation 20.
\begin{equation}
    \mathit{L} = \mathit{L}_{RL} + \mathit{L}_{decision} + \mathit{L}_{feature}
\end{equation}
Where $\mathit{L}_{RL}$ is the training loss defined in PPO or DQN, $\mathit{L}_{decision}$ is the decision loss and $\mathit{L}_{feature}$ is the feature loss.


From above, some details about our OPD-DA pipeline are shown in Algorithm 1.

\begin{algorithm}[tb]
\caption{Online Policy Distillation with Decision-Attention}
\label{alg:algorithm}
\textbf{Input}: Policies : $\left\{ A \right\}_{i=1}^{T} $\\
\textbf{Parameter}: Epoch number : $epoch$ \\
\textbf{Output}: Trained Policies : $\left\{ A \right\}_{i=1}^{T} $

\begin{algorithmic}[1] 
\FOR{$p=1$; $p \leq epoch$; $p+=1$}
\WHILE{Not End}
\STATE Given state : $S_{j}$
\FOR{$i=1$; $i \leq T$; $i+=1$}

\STATE Get output and features of policy $i$ :  $O_{i}^{j}, F_{i}^{j} = A_{i}(S_{j})$
\STATE Decision-Attention weights : \\ $W_{q}^{j} = softmax(\frac{ O_{i}^{j} O_{q}^{j} }{\sqrt{d_{k}}})O_{q}^{j} $, where $q \in [1,T] $ and $ q \neq i$
\STATE Get decision-based supervisory information : $ \mathit{H}_{decision}^{i} = \sum W_{q}^{j} * O_{q}^{j} $ where $q \in [1,T] $ and $ q \neq i$
\STATE Get feature-based supervisory information : $ \mathit{H}_{feature}^{i} = \sum W_{q}^{j} * F_{q}^{j} $ where $q \in [1,T] $ and $ q \neq i$
\STATE Get RL loss : $\mathit{L}_{RL}^{i}$
\STATE Get decision loss : $\mathit{L}_{decision}^{i} = \mathit{L}(O_{i}^{j}, \mathit{H}_{decision}^{i})$
\STATE Get feature loss : $\mathit{L}_{feature}^{i} = \mathit{L}(F_{i}^{j}, \mathit{H}_{feature}^{i})$
\STATE Get loss function : $\mathit{L}^{i} = \mathit{L}_{RL}^{i} + \mathit{L}_{decision}^{i} + \mathit{L}_{feature}^{i} $
\STATE Update policy : $\left\{ A \right\}_{i} $
\ENDFOR
\STATE Given new state : $S_{j}$
\ENDWHILE



\ENDFOR
\end{algorithmic}
\end{algorithm}
\section{Experiments}
\begin{figure}[t]
    \centering
    \includegraphics[width=\linewidth]{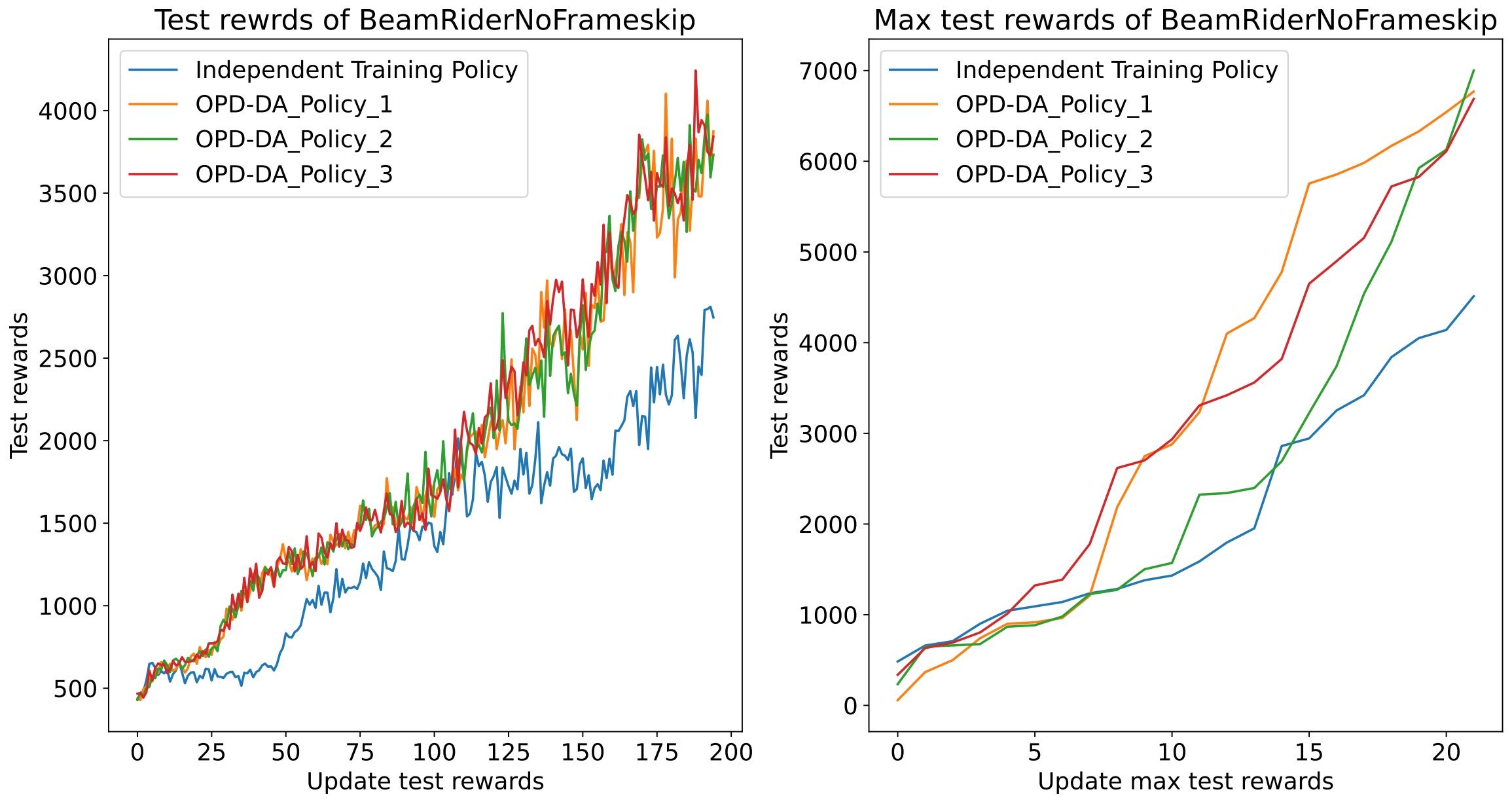}
    \caption{Results of BeamRiderNoFrameskip on PPO.}
    \label{fig:enter-label}
\end{figure}

\begin{figure}[t]
    \centering
    \includegraphics[width=\linewidth]{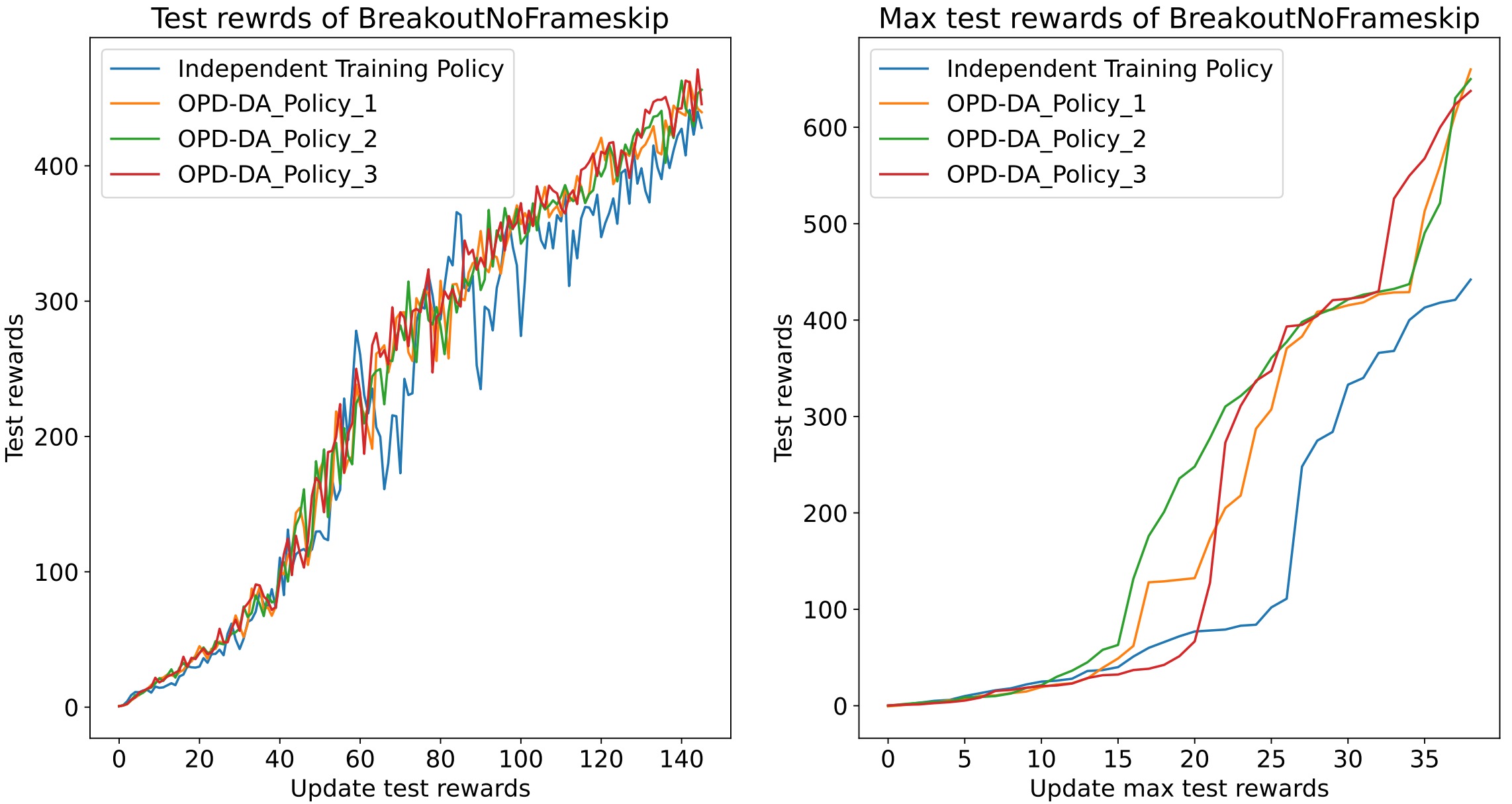}
    \caption{Results of BreakoutNoFrameskip on PPO.}
    \label{fig:enter-label}
\end{figure}

\begin{figure}[t]
    \centering
    \includegraphics[width=\linewidth]{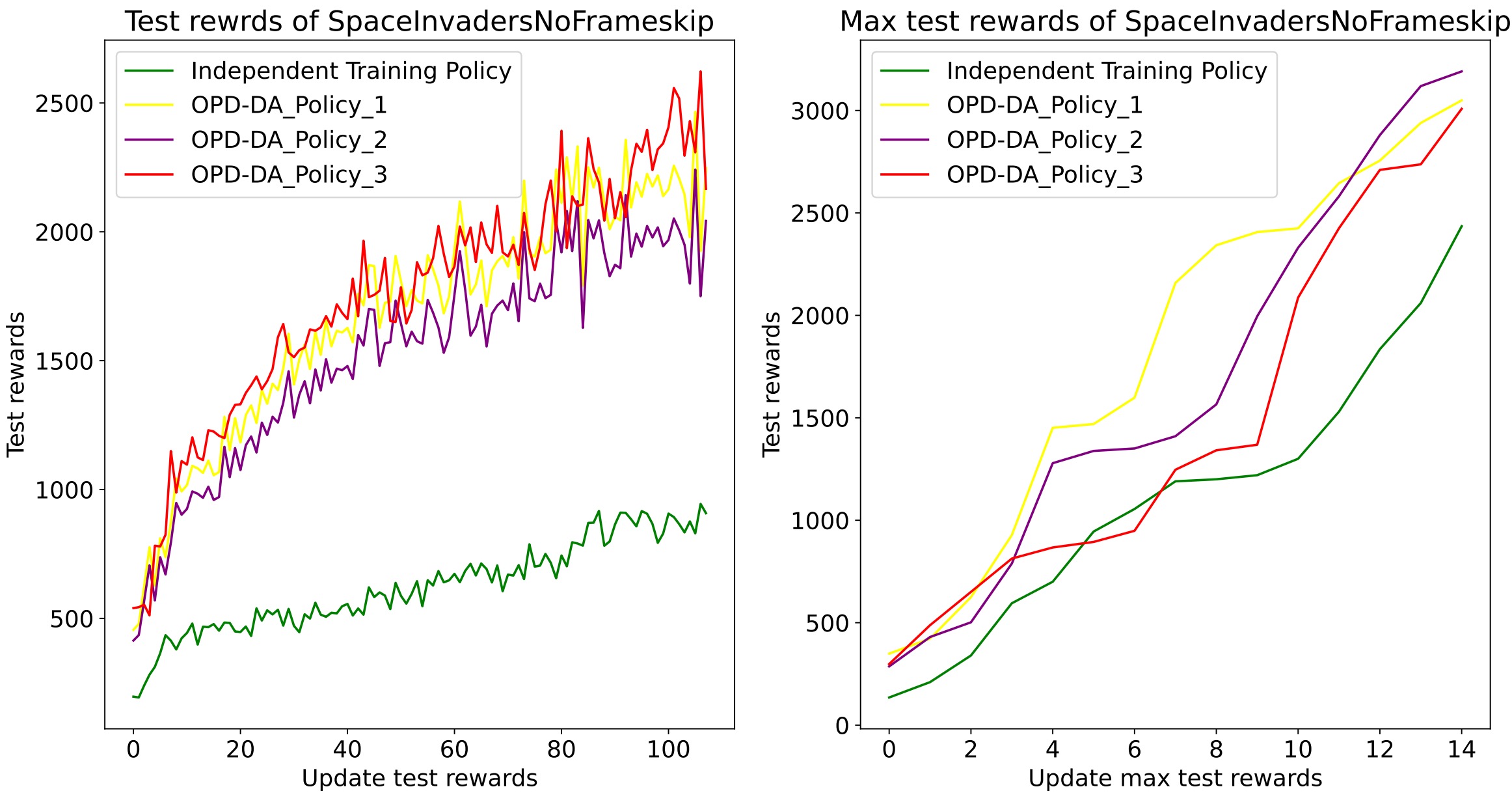}
    \caption{Results of SpaceInvaderNoFrameskip on DQN.}
    \label{fig:enter-label}
\end{figure}

\begin{figure}[t]
    \centering
    \includegraphics[width=\linewidth]{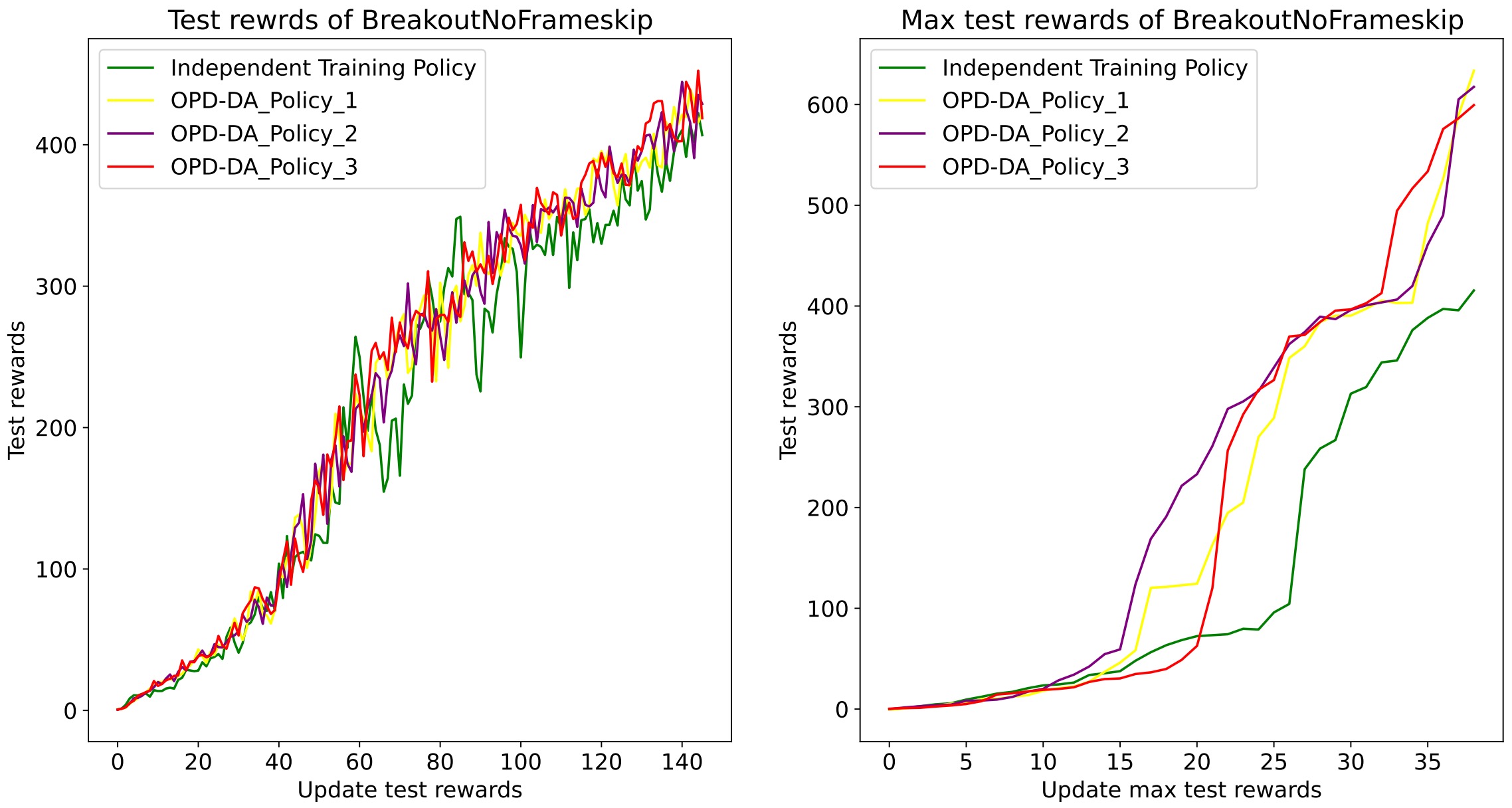} 
    \caption{Results of BreakoutNoFrameskip on DQN.}
    \label{fig:enter-label}
\end{figure}


\begin{figure}[t]
    \centering
    \includegraphics[width=\linewidth]{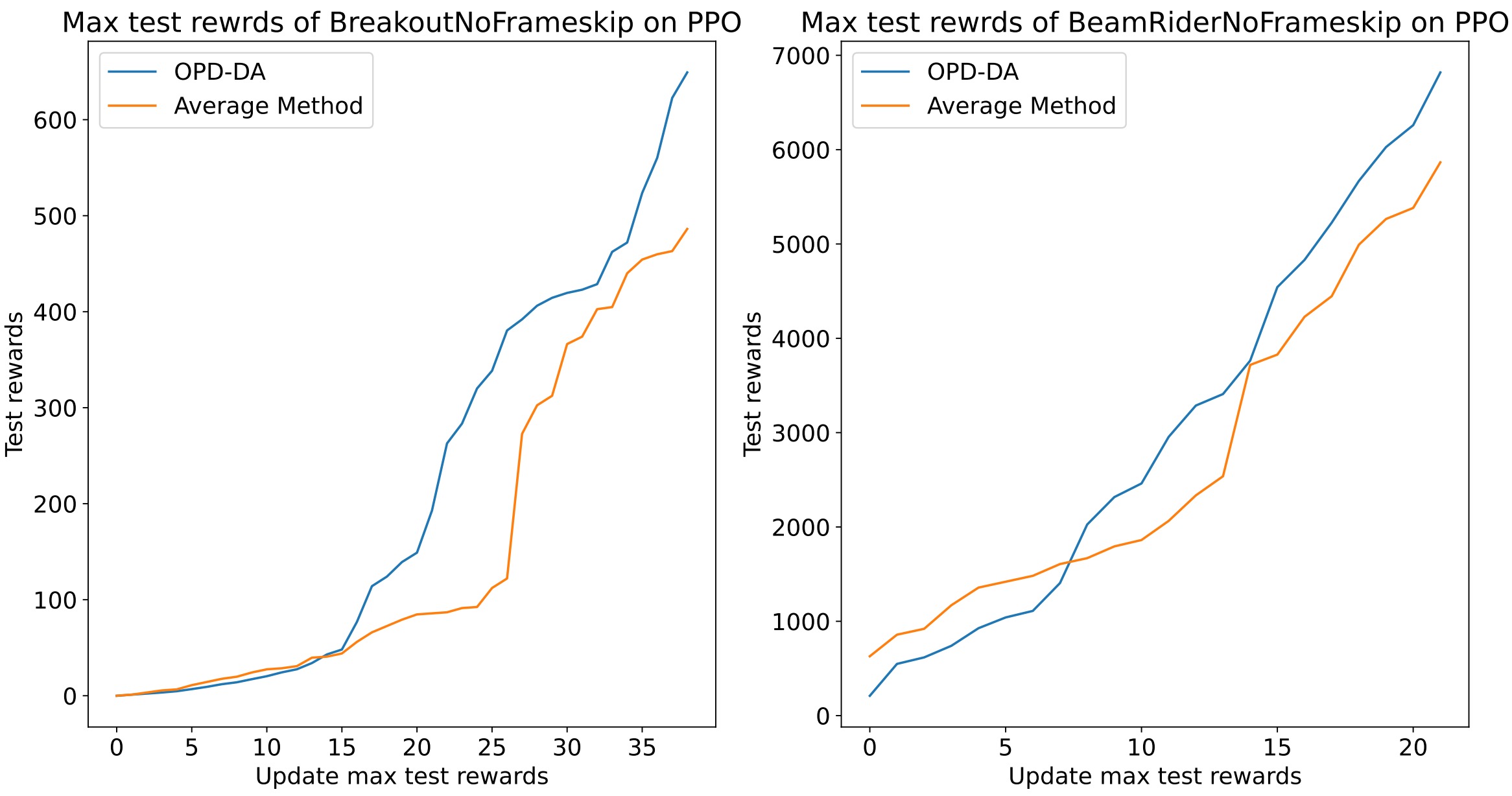}
    \caption{Comparing with policy distillation using same weights, in both methods, we use three policies and PPO algorithm for training on different tasks. In this figure, we give the average rewards of three policies in both methods.}
    \label{fig:enter-label}
\end{figure}

\begin{figure}[t]
    \centering
    \includegraphics[width=\linewidth]{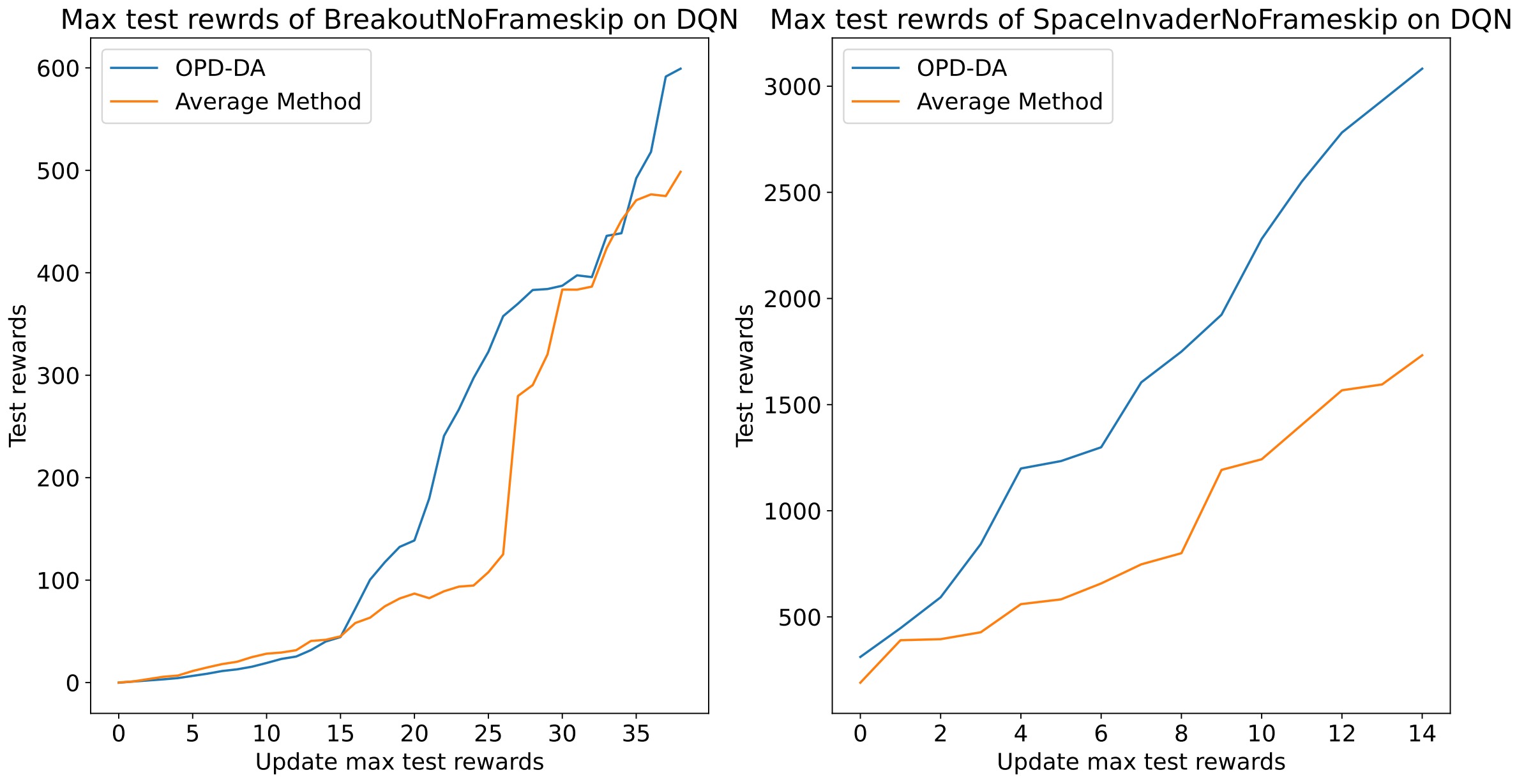}
    \caption{Comparing with policy distillation using same weights, in both methods, we use three policies and DQN algorithm for training on different tasks. In this figure, we give the average rewards of three policies in both methods.}
    \label{fig:enter-label}
\end{figure}

\begin{figure}[t]
    \centering
     \includegraphics[width=\linewidth]{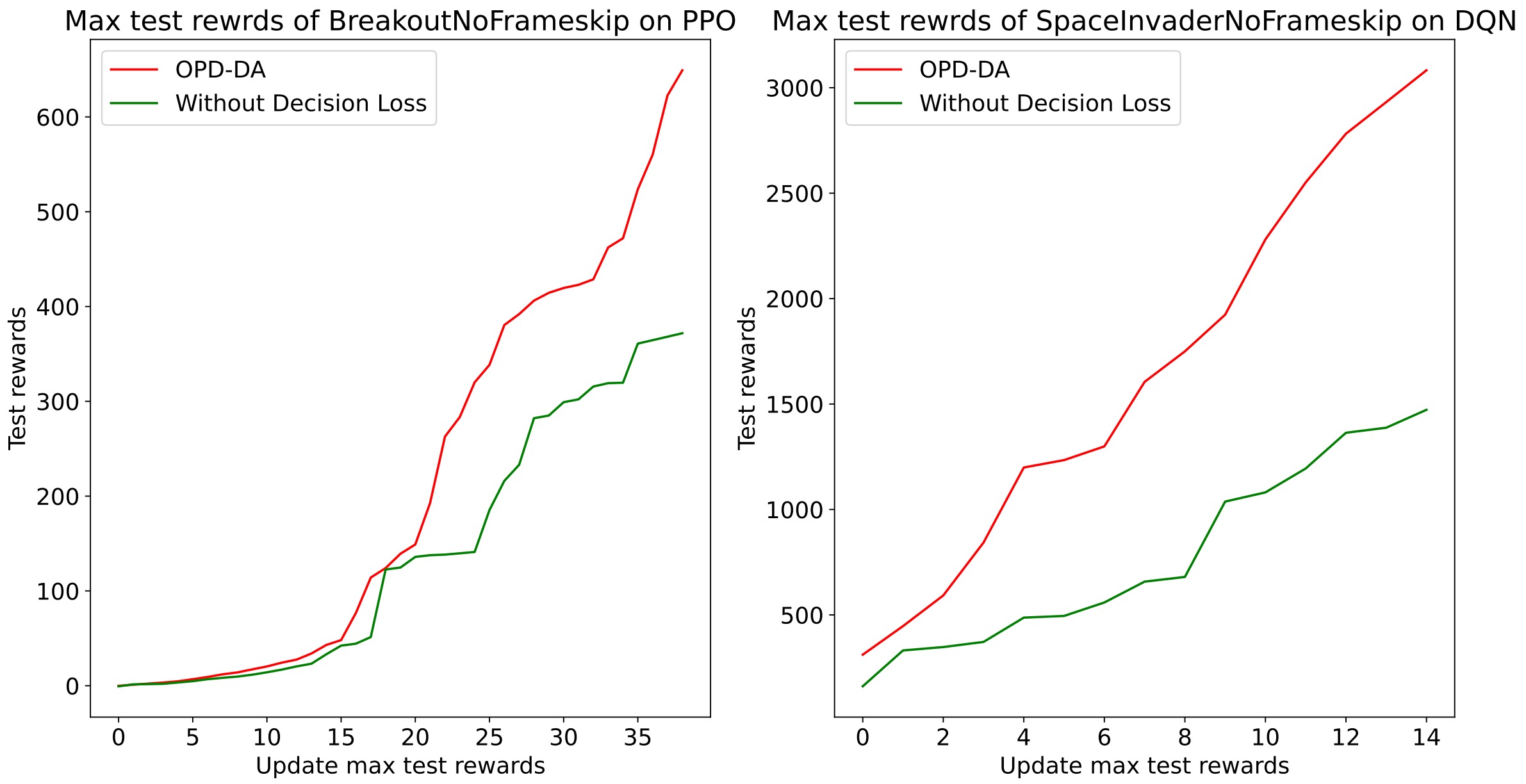}
    \caption{In the ablation experiments, the average performance of three policies for our method and online policy distillation without decision loss.}
    \label{fig:enter-label}
\end{figure}

\begin{figure}[t]
    \centering
    \includegraphics[width=\linewidth]{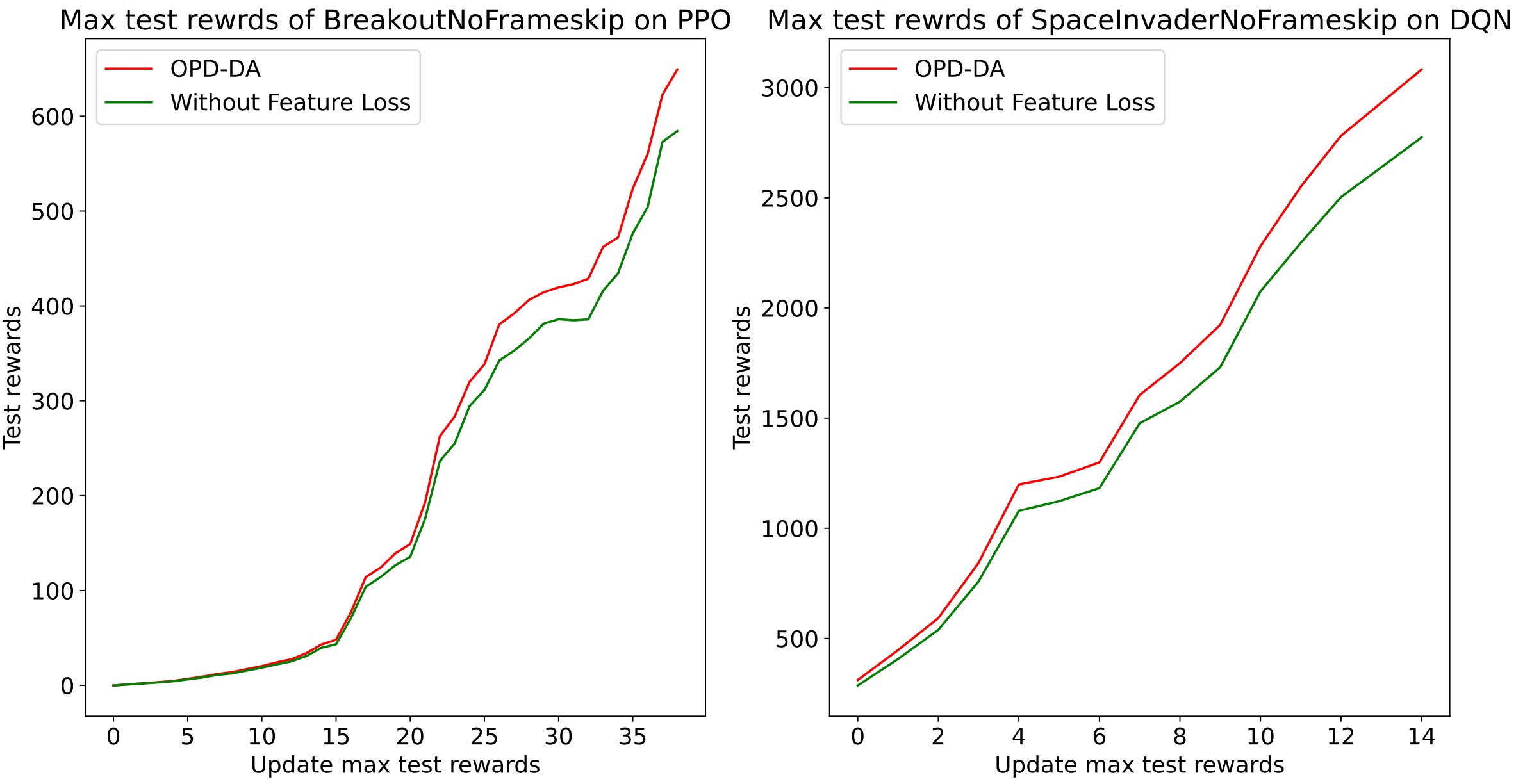}
    \caption{In the ablation experiments, the average performance of three policies for our method and online policy distillation without feature loss.}
    \label{fig:enter-label}
\end{figure}

  
\begin{table}
\centering
  \caption{For PPO algorithm, the rewards obtained by the policies on different tasks.}
  \label{tab:sample}
  \begin{tabular}{ccc}
   \hline
    & \textbf{Breakout} & \textbf{BeamRider}  \\
\hline
    Agent & 442 & 4512 \\
    Online\_Policy\_1 & \textbf{660} ($\uparrow$ \textbf{49.3\%}) & 6768 ($\uparrow$ 50.0\%)\\
    Online\_Policy\_2 & 650 ($\uparrow$ 47.1\%) & \textbf{7000} ($\uparrow$ \textbf{55.1\%})\\
    Online\_Policy\_3 & 637 ($\uparrow$ 44.1\%) & 6688 ($\uparrow$ 48.2\%)\\
\hline
\end{tabular}
\end{table}

\begin{table}
  \centering
\caption{For DQN algorithm, the rewards obtained by the policies on different tasks.}

  \begin{tabular}{c|c|c}
\hline

     & \textbf{SpaceInvader} &   \textbf{Breakout}\\
\hline
    Agent & 2435  & 430 \\
    Online\_Policy\_1 & 3161 ($\uparrow$ 29.8\%) & \textbf{620} ($\uparrow$ \textbf{44.2\%}) \\ 
    Online\_Policy\_2 & \textbf{3401}  ($\uparrow$ \textbf{39.7\%}) & 610 ($\uparrow$ 41.9\%) \\
    Online\_Policy\_3 &  3039 ($\uparrow$ 24.8\%) & 590 ($\uparrow$ 37.2\%) \\
\hline

  \end{tabular}
  
  \label{tab:sample}
\end{table}

\subsection{Experiment Platform}
The Atari game platform has played an important role in the field of reinforcement learning, serving as the basis for many studies and experiments \cite{lehmkuhl2019deep}. This is mainly due to a study called "DQN" (Deep Q-Network) released by OpenAI in 2013, which used Atari games as a test environment and demonstrated the potential of deep reinforcement learning on game tasks. The Atari game platform contains a variety of games that can be used as a test environment for reinforcement learning algorithms. The Atari game platform continues to serve as a benchmarking environment and research tool in the field of reinforcement learning, with many researchers conducting experiments on this platform and exploring various algorithms and techniques to improve performance on gaming tasks. 
In our experiments, we choose different games and use PPO and DQN algorithms to solve these tasks.
\subsection{Algorithms}
In reinforcement learning, there are many different algorithms.
Proximal Policy Optimization (PPO) and Deep Q-Network (DQN) are two important algorithms in reinforcement learning.
Thus, in our experiments, we use PPO and DQN to solve tasks in Atari games. 

\textbf{PPO.} PPO belongs to the class of policy optimization methods and is known for its stability and strong empirical performance. And PPO can tackle the high variance problem in policy gradient methods by introducing a clipped objective function.
In PPO, the Actor and Critic are two fundamental components that work together to optimize the policy and estimate the value function. In our experiments, the actor and critic share the common backbone, which includes some CNN layers and extracts feature of inputs.

\textbf{DQN.} DQN takes the state as input and outputs Q-values for each possible action by convolutional neural networks. DQN uses replay memory to break the temporal correlations in the data, making training more stable and efficient.
DQN uses two separate neural networks, the "online" network and the "target" network. The online network is used to select actions, and the target network stabilizes training by providing more stable Q-value targets during learning.


\subsection{Results of Our Method}
In our experiments, we use three policies for online distillation. In Figure 2, 3, 4 and 5, each Figure consists of left and right parts. In the left part, we show test rewards curve during training, and smooth the curve using the average of 50 consecutive rewards. In the right part, we show the update curve of the maximum test rewards during training.


\textbf{PPO.} Figure 2 and 3 show the test rewards of these agents during training on BeamRiderNoFrameskip and BreakoutNoFrameskip. Comparing to an independent training agent, Table 1 shows our OPD-DA can improve policy performance by 46\% on average for BreakoutNoFrameskip and 51\% on average for BeamRiderNoFrameskip. 



\textbf{DQN.} Figure 4 and 5 show the test rewards of these policies during training on SpaceInvaderNoFrameskip and BreakoutNoFrameskip. Comparing to an independent training policy, Table 2 shows our OPD-DA can improve policy performance by 31\% on average for SpaceInvaderNoFrameskip and 41\% on average for BreakoutNoFrameskip. 


The above experimental results show that our OPD-DA framework can make a policy learn well from others, and make these policies perform better on both PPO and DQN algorithms for different tasks.

\subsection{Ablation Study}

In our method, we use the response and intermediate layer features of each policy as supervisory information. 
To prove both supervisory information is valid, we make ablation experiments under two settings including without decision loss and feature loss. 

\textbf{Decision Loss} 
To verify the effectiveness of OPD-DA and the decision loss on making one policy learn from others' output. We make the ablation experiments by comparing our method with the online policy distillation method without decision loss. Figure 8 shows our decision loss can make OPD-DA perform better on both PPO and DQN algorithms. Specifically, when using the DQN algorithm to solve SpaceInvaderNoFrameskip task, the reward obtained by OPD-DA is 93.75\% higher than the method without decision loss. When using the PPO algorithm to solve BreakoutNoFrameskip task, the reward obtained by OPD-DA is 66.67\% higher.

\textbf{Feature Loss} 
Besides the decision loss, we also verify the effectiveness of the feature loss which can make one policy learn from others' intermediate layer features. We make the ablation experiments by comparing our OPD-DA framework with the online policy distillation method without feature loss. Figure 9 shows our feature loss can make OPD-DA  perform better on both PPO and DQN algorithms. Specifically, when using the DQN algorithm to solve SpaceInvaderNoFrameskip task, the reward obtained by OPD-DA is 6.90\% higher than the method without decision loss. When using the PPO algorithm to solve BreakoutNoFrameskip task, the reward obtained by OPD-DA is 8.33\% higher.

\textbf{Decision-Attention} 
To verify our Decision-Attention module can highlight differences between different decisions on both PPO and DQN algorithms.
We make the experiments by comparing with the method of averaging expected rewards as a supervision signal.
In Figure 6 and Figure 7, we give the average rewards of the three policies on different policy distillation methods, including our OPD-DA method and method of weighting supervisory information using equal weights.

For PPO, we make these ablation experiments on BreakoutNoFramkeskip and BeamRiderNoFrameskip to validate the effectiveness of our Decision-Attention Module. Figure 6 shows that our method can obtain more test rewards on both tasks. For BreakoutNoFramkeskip and BeamRiderNoFrameskip, the reward received by using our Decision-Attention module is 35.42\% and 16.67\% higher than the policy distillation method using the same weights.

For DQN, we make these ablation experiments on BreakoutNoFrameskip and SpaceInvaderNoFrameskip to validate that our Decision-Attention module is effective on the DQN algorithm. Figure 7 shows that our method can make policies perform better on both tasks. For BreakoutNoFrameskip and SpaceInvaderNoFrameskip, the reward obtained by using our Decision-Attention module is 26.53\% and 72.2\% higher than the policy distillation method without the attention mechanism.

\section{Conclusion}
Online policy distillation is an excellent way to transfer knowledge when a pre-trained, high-performance teacher policy is not easily accessible.
We propose a novel online policy distillation framework that makes policies learn from each other. 
However, naive aggregation functions tend to cause student policies to homogenize quickly. Thus, we propose the Decision-Attention module to assign various weights to group members. Experimental evidence proves the superiority of the Decision-Attention module, which makes our OPD-DA framework perform better.

\section*{Acknowledgement}
This work is partially supported by the Chinese Academy of Sciences Specific Research Assistant Funding Project and the Beijing Natural Science Foundation under grant 4244098. 

\bibliographystyle{IEEEtran}
\bibliography{IEEEabrv,references}

\begin{thebibliography}{10}
\providecommand{\url}[1]{#1}
\csname url@samestyle\endcsname
\providecommand{\newblock}{\relax}
\providecommand{\bibinfo}[2]{#2}
\providecommand{\BIBentrySTDinterwordspacing}{\spaceskip=0pt\relax}
\providecommand{\BIBentryALTinterwordstretchfactor}{4}
\providecommand{\BIBentryALTinterwordspacing}{\spaceskip=\fontdimen2\font plus
\BIBentryALTinterwordstretchfactor\fontdimen3\font minus \fontdimen4\font\relax}
\providecommand{\BIBforeignlanguage}[2]{{%
\expandafter\ifx\csname l@#1\endcsname\relax
\typeout{** WARNING: IEEEtran.bst: No hyphenation pattern has been}%
\typeout{** loaded for the language `#1'. Using the pattern for}%
\typeout{** the default language instead.}%
\else
\language=\csname l@#1\endcsname
\fi
#2}}
\providecommand{\BIBdecl}{\relax}
\BIBdecl

\bibitem{li2017deep}
Y.~Li, ``Deep reinforcement learning: An overview,'' \emph{arXiv preprint arXiv:1701.07274}, 2017.

\bibitem{rusu2015policy}
A.~A. Rusu, S.~G. Colmenarejo, C.~Gulcehre, G.~Desjardins, J.~Kirkpatrick, R.~Pascanu, V.~Mnih, K.~Kavukcuoglu, and R.~Hadsell, ``Policy distillation,'' \emph{arXiv preprint arXiv:1511.06295}, 2015.

\bibitem{lai2020dual}
K.-H. Lai, D.~Zha, Y.~Li, and X.~Hu, ``Dual policy distillation,'' \emph{arXiv preprint arXiv:2006.04061}, 2020.

\bibitem{romero2014fitnets}
A.~Romero, N.~Ballas, S.~E. Kahou, A.~Chassang, C.~Gatta, and Y.~Bengio, ``Fitnets: Hints for thin deep nets,'' \emph{arXiv preprint arXiv:1412.6550}, 2014.

\bibitem{yang2023categories}
C.~Yang, X.~Yu, Z.~An, and Y.~Xu, ``Categories of response-based, feature-based, and relation-based knowledge distillation,'' in \emph{Advancements in Knowledge Distillation: Towards New Horizons of Intelligent Systems}.\hskip 1em plus 0.5em minus 0.4em\relax Springer, 2023, pp. 1--32.

\bibitem{lu2019artificial}
Y.~Lu, ``Artificial intelligence: a survey on evolution, models, applications and future trends,'' \emph{Journal of Management Analytics}, vol.~6, no.~1, pp. 1--29, 2019.

\bibitem{liu2020new}
H.~Liu, C.~Yu, H.~Wu, Z.~Duan, and G.~Yan, ``A new hybrid ensemble deep reinforcement learning model for wind speed short term forecasting,'' \emph{Energy}, vol. 202, p. 117794, 2020.

\bibitem{radford2023robust}
A.~Radford, J.~W. Kim, T.~Xu, G.~Brockman, C.~McLeavey, and I.~Sutskever, ``Robust speech recognition via large-scale weak supervision,'' in \emph{International Conference on Machine Learning}.\hskip 1em plus 0.5em minus 0.4em\relax PMLR, 2023, pp. 28\,492--28\,518.

\bibitem{dong2021survey}
S.~Dong, P.~Wang, and K.~Abbas, ``A survey on deep learning and its applications,'' \emph{Computer Science Review}, vol.~40, p. 100379, 2021.

\bibitem{mnih2013playing}
V.~Mnih, K.~Kavukcuoglu, D.~Silver, A.~Graves, I.~Antonoglou, D.~Wierstra, and M.~Riedmiller, ``Playing atari with deep reinforcement learning,'' \emph{arXiv preprint arXiv:1312.5602}, 2013.

\bibitem{schulman2017proximal}
J.~Schulman, F.~Wolski, P.~Dhariwal, A.~Radford, and O.~Klimov, ``Proximal policy optimization algorithms,'' \emph{arXiv preprint arXiv:1707.06347}, 2017.

\bibitem{zhang2018deep}
Y.~Zhang, T.~Xiang, T.~M. Hospedales, and H.~Lu, ``Deep mutual learning,'' in \emph{Proceedings of the IEEE conference on computer vision and pattern recognition}, 2018, pp. 4320--4328.

\bibitem{Gong_Lin_Zhang_Shen_Li_Qiao_Ren_Li_Yu_Ma_2023}
\BIBentryALTinterwordspacing
L.~Gong, S.~Lin, B.~Zhang, Y.~Shen, K.~Li, R.~Qiao, B.~Ren, M.~Li, Z.~Yu, and L.~Ma, ``\BIBforeignlanguage{en-US}{Adaptive hierarchy-branch fusion for online knowledge distillation},'' \emph{\BIBforeignlanguage{en-US}{Proceedings of the AAAI Conference on Artificial Intelligence}}, vol.~37, no.~6, p. 7731–7739, Jun 2023. [Online]. Available: \url{http://dx.doi.org/10.1609/aaai.v37i6.25937}
\BIBentrySTDinterwordspacing

\bibitem{chen2020online}
D.~Chen, J.-P. Mei, C.~Wang, Y.~Feng, and C.~Chen, ``Online knowledge distillation with diverse peers,'' in \emph{Proceedings of the AAAI conference on artificial intelligence}, vol.~34, no.~04, 2020, pp. 3430--3437.

\bibitem{zhu2018knowledge}
X.~Zhu, S.~Gong \emph{et~al.}, ``Knowledge distillation by on-the-fly native ensemble,'' \emph{Advances in neural information processing systems}, vol.~31, 2018.

\bibitem{yang2021multi}
C.~Yang, Z.~An, and Y.~Xu, ``Multi-view contrastive learning for online knowledge distillation,'' in \emph{ICASSP 2021-2021 IEEE International Conference on Acoustics, Speech and Signal Processing (ICASSP)}.\hskip 1em plus 0.5em minus 0.4em\relax IEEE, 2021, pp. 3750--3754.

\bibitem{yang2022mutual}
C.~Yang, Z.~An, L.~Cai, and Y.~Xu, ``Mutual contrastive learning for visual representation learning,'' in \emph{Proceedings of the AAAI Conference on Artificial Intelligence}, vol.~36, no.~3, 2022, pp. 3045--3053.

\bibitem{yang2023online}
C.~Yang, Z.~An, H.~Zhou, F.~Zhuang, Y.~Xu, and Q.~Zhang, ``Online knowledge distillation via mutual contrastive learning for visual recognition,'' \emph{IEEE Transactions on Pattern Analysis and Machine Intelligence}, vol.~45, no.~8, pp. 10\,212--10\,227, 2023.

\bibitem{huang2024kfc}
L.~Huang, Z.~An, Y.~Zeng, Y.~Xu \emph{et~al.}, ``Kfc: Knowledge reconstruction and feedback consolidation enable efficient and effective continual generative learning,'' in \emph{The Second Tiny Papers Track at ICLR 2024}.

\bibitem{hinton2015distilling}
G.~Hinton, O.~Vinyals, and J.~Dean, ``Distilling the knowledge in a neural network,'' \emph{arXiv preprint arXiv:1503.02531}, 2015.

\bibitem{yang2021hierarchical}
C.~Yang, Z.~An, L.~Cai, and Y.~Xu, ``Hierarchical self-supervised augmented knowledge distillation,'' in \emph{Proceedings of the Thirtieth International Joint Conference on Artificial Intelligence (IJCAI)}, 2021, pp. 1217--1223.

\bibitem{mirzadeh2020improved}
S.~I. Mirzadeh, M.~Farajtabar, A.~Li, N.~Levine, A.~Matsukawa, and H.~Ghasemzadeh, ``Improved knowledge distillation via teacher assistant,'' in \emph{Proceedings of the AAAI conference on artificial intelligence}, vol.~34, no.~04, 2020, pp. 5191--5198.

\bibitem{zhang2020prime}
Y.~Zhang, Z.~Lan, Y.~Dai, F.~Zeng, Y.~Bai, J.~Chang, and Y.~Wei, ``Prime-aware adaptive distillation,'' in \emph{Computer Vision--ECCV 2020: 16th European Conference, Glasgow, UK, August 23--28, 2020, Proceedings, Part XIX 16}.\hskip 1em plus 0.5em minus 0.4em\relax Springer, 2020, pp. 658--674.

\bibitem{yang2022mixskd}
C.~Yang, Z.~An, H.~Zhou, L.~Cai, X.~Zhi, J.~Wu, Y.~Xu, and Q.~Zhang, ``Mixskd: Self-knowledge distillation from mixup for image recognition,'' in \emph{European Conference on Computer Vision}.\hskip 1em plus 0.5em minus 0.4em\relax Springer, 2022, pp. 534--551.

\bibitem{huang2024etag}
L.~Huang, Y.~Zeng, C.~Yang, Z.~An, B.~Diao, and Y.~Xu, ``etag: Class-incremental learning via embedding distillation and task-oriented generation,'' in \emph{Proceedings of the AAAI Conference on Artificial Intelligence}, vol.~38, no.~11, 2024, pp. 12\,591--12\,599.

\bibitem{huang2017like}
Z.~Huang and N.~Wang, ``Like what you like: Knowledge distill via neuron selectivity transfer,'' \emph{arXiv preprint arXiv:1707.01219}, 2017.

\bibitem{yang2022knowledge}
C.~Yang, Z.~An, L.~Cai, and Y.~Xu, ``Knowledge distillation using hierarchical self-supervision augmented distribution,'' \emph{IEEE Transactions on Neural Networks and Learning Systems}, vol.~35, no.~2, pp. 2094--2108, 2024.

\bibitem{yang2022cross}
C.~Yang, H.~Zhou, Z.~An, X.~Jiang, Y.~Xu, and Q.~Zhang, ``Cross-image relational knowledge distillation for semantic segmentation,'' in \emph{Proceedings of the IEEE/CVF Conference on Computer Vision and Pattern Recognition}, 2022, pp. 12\,319--12\,328.

\bibitem{yu2023dsformer}
C.~Yu, F.~Wang, Z.~Shao, T.~Sun, L.~Wu, and Y.~Xu, ``Dsformer: A double sampling transformer for multivariate time series long-term prediction,'' in \emph{Proceedings of the 32nd ACM International Conference on Information and Knowledge Management}, 2023, pp. 3062--3072.

\bibitem{yu2024ginar}
C.~Yu, F.~Wang, Z.~Shao, T.~Qian, Z.~Zhang, W.~Wei, and Y.~Xu, ``Ginar: An end-to-end multivariate time series forecasting model suitable for variable missing,'' \emph{arXiv preprint arXiv:2405.11333}, 2024.

\bibitem{vaswani2017attention}
A.~Vaswani, N.~Shazeer, N.~Parmar, J.~Uszkoreit, L.~Jones, A.~N. Gomez, {\L}.~Kaiser, and I.~Polosukhin, ``Attention is all you need,'' \emph{Advances in neural information processing systems}, vol.~30, 2017.

\bibitem{han2022survey}
K.~Han, Y.~Wang, H.~Chen, X.~Chen, J.~Guo, Z.~Liu, Y.~Tang, A.~Xiao, C.~Xu, Y.~Xu \emph{et~al.}, ``A survey on vision transformer,'' \emph{IEEE transactions on pattern analysis and machine intelligence}, vol.~45, no.~1, pp. 87--110, 2022.

\bibitem{yang2020gated}
C.~Yang, Z.~An, H.~Zhu, X.~Hu, K.~Zhang, K.~Xu, C.~Li, and Y.~Xu, ``Gated convolutional networks with hybrid connectivity for image classification,'' in \emph{Proceedings of the AAAI Conference on Artificial Intelligence}, vol.~34, no.~07, 2020, pp. 12\,581--12\,588.

\bibitem{devlin2018bert}
J.~Devlin, M.-W. Chang, K.~Lee, and K.~Toutanova, ``Bert: Pre-training of deep bidirectional transformers for language understanding,'' \emph{arXiv preprint arXiv:1810.04805}, 2018.

\bibitem{radford2018improving}
A.~Radford, K.~Narasimhan, T.~Salimans, I.~Sutskever \emph{et~al.}, ``Improving language understanding by generative pre-training,'' 2018.

\bibitem{li2023knowledge}
L.~Li, W.~Su, F.~Liu, M.~He, and X.~Liang, ``Knowledge fusion distillation: Improving distillation with multi-scale attention mechanisms,'' \emph{Neural Processing Letters}, pp. 1--16, 2023.

\bibitem{wiering2012reinforcement}
M.~A. Wiering and M.~Van~Otterlo, ``Reinforcement learning,'' \emph{Adaptation, learning, and optimization}, vol.~12, no.~3, p. 729, 2012.

\bibitem{fukuda2017efficient}
T.~Fukuda, M.~Suzuki, G.~Kurata, S.~Thomas, J.~Cui, and B.~Ramabhadran, ``Efficient knowledge distillation from an ensemble of teachers.'' in \emph{Interspeech}, 2017, pp. 3697--3701.

\bibitem{sutton2018reinforcement}
R.~S. Sutton and A.~G. Barto, \emph{Reinforcement learning: An introduction}.\hskip 1em plus 0.5em minus 0.4em\relax MIT press, 2018.

\bibitem{lehmkuhl2019deep}
J.~Lehmkuhl and D.~Bick, ``Deep reinforcement learning in atari 2600 games,'' Ph.D. dissertation, 2019.

\end{thebibliography}
\end{document}